\documentclass[letterpaper, 10 pt, conference]{ieeeconf}  % Comment this line out if you need a4paper

\IEEEoverridecommandlockouts                              % This command is only needed if 
\usepackage{graphics} % for pdf, bitmapped graphics files
\graphicspath{ {./images/} }
\usepackage{subcaption}

\usepackage{amsmath} % assumes amsmath package installed
\DeclareMathOperator*{\argmax}{argmax}
\usepackage{amssymb}  % assumes amsmath package installed
\usepackage{algorithm}
\usepackage[noend]{algpseudocode}
\usepackage{lipsum}
\usepackage{mwe}
\usepackage{gensymb}
\usepackage{amsthm}
\usepackage{float}
\usepackage{booktabs}
\usepackage{multirow}
\let\labelindent\relax
\usepackage{enumitem} % leftmargin for itemize
\usepackage[dvipsnames]{xcolor} % for color text
\usepackage{soul} % for underline
\setul{0.3ex}{0.2ex}

\usepackage[breaklinks,colorlinks]{hyperref}
\newtheoremstyle{mystyle}%                % Name
  {}%                                     % Space above
  {}%                                     % Space below
  {\itshape}%                                     % Body font
  {}%                                     % Indent amount
  {\bfseries}%                            % Theorem head font
  {.}%                                    % Punctuation after theorem head
  { }%                                    % Space after theorem head, ' ', or \newline
  {}%                                     % Theorem head spec (can be left empty, meaning `normal')
\theoremstyle{mystyle}
\newtheorem{definition}{Definition}

\newtheorem{assumption}{Assumption}
\usepackage{todonotes} % for missingfigure

\title{\huge
Adversarial Object Rearrangement in Constrained Environments with Heterogeneous Graph Neural Networks
}
\author{Xibai Lou$^{1}$, Houjian Yu$^{1}$, Ross Worobel$^{2}$, Yang Yang$^{2}$ and Changhyun Choi$^{1}$\\
\thanks{*This work was in part supported by UMII MnDRIVE Ph.D. Graduate Assistantship and MnDRIVE Initiative on Robotics, Sensors, and Advanced Manufacturing.}% <-this % stops a space
\thanks{$^{1}$ The authors are with the Department of Electrical and Computer Engineering, Univ. of Minnesota, Minneapolis, USA
        {\tt\small \{lou00015, yu000487, cchoi\}@umn.edu}}
\thanks{$^{2}$ The authors are with the Department of Computer Science and Engineering, Univ. of Minnesota, Minneapolis, USA
        {\tt\small \{worob006, yang5276\}@umn.edu}}
}

\begin{document}

\maketitle
\thispagestyle{empty}
\pagestyle{empty}

%%%%%%%%%%%%%%%%%%%%%%%%%%%%%%%%%%%%%%%%%%%%%%%%%%%%%%%%%%%%%%%%%%%%%%%%%%%%%%%%
\begin{abstract}
Adversarial object rearrangement in the real world (e.g., previously unseen or oversized items in kitchens and stores) could benefit from understanding task scenes, which inherently entail heterogeneous components such as current objects, goal objects, and environmental constraints. The semantic relationships among these components are distinct from each other and crucial for multi-skilled robots to perform efficiently in everyday scenarios. We propose a hierarchical robotic manipulation system that learns the underlying relationships and maximizes the collaborative power of its diverse skills (e.g., \textsc{pick-place}, \textsc{push}) for rearranging adversarial objects in constrained environments. The high-level coordinator employs a heterogeneous graph neural network (HetGNN), which reasons about the current objects, goal objects, and environmental constraints; the low-level 3D Convolutional Neural Network-based actors execute the action primitives. Our approach is trained entirely in simulation, and achieved an average success rate of 87.88\% and a planning cost of 12.82 in real-world experiments, surpassing all baseline methods. Supplementary material is available at \url{https://sites.google.com/umn.edu/versatile-rearrangement}.
\end{abstract}
\smallbreak

%%%%%%%%%%%%%%%%%%%%%%%%%%%%%%%%%%%%%%%%%%%%%%%%%%%%%%%%%%%%%%%%%%%%%%%%%%%%%%%%
\section{INTRODUCTION}
\begin{figure}[t]
    \includegraphics[width=0.485\textwidth]{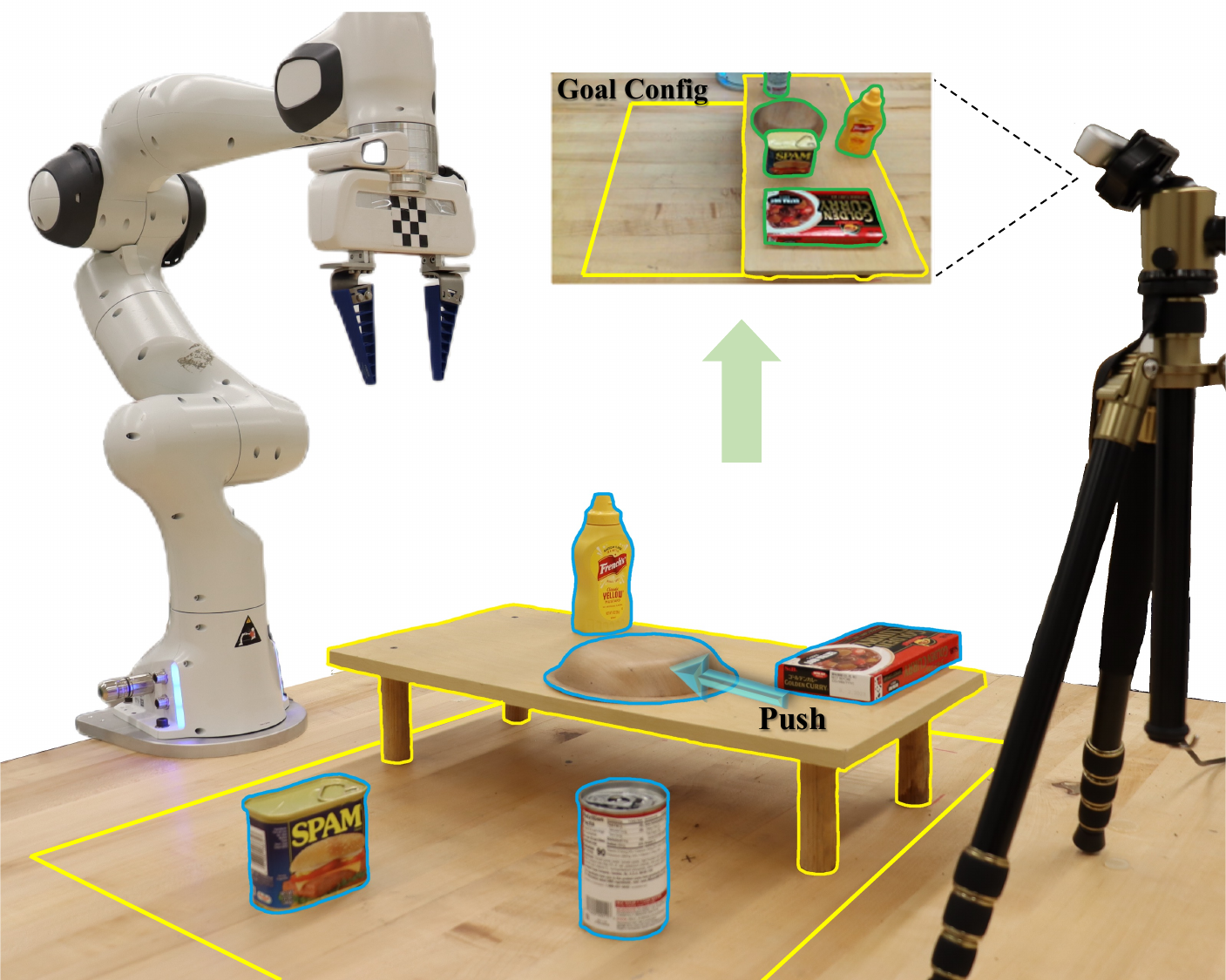}
    \caption{In this adversarial object rearrangement task, the color-coded heterogeneous task components (e.g., \setulcolor{cyan}\ul{current objects}, \setulcolor{Green}\ul{goal objects}, and \setulcolor{Goldenrod}\ul{environmental constraints}) are linked by different semantic relationships that are crucial to efficiently guiding a multi-skilled robot. By understanding that the ``bowl'' and the ``shelf'' are related by ``on'', a robot will swiftly push it to the nearby goal and clear space for the ``meat can'', which requires \textsc{pick-place} to move from ``ground'' to ``shelf''. }
  \label{fig:cover}
  \vspace{-8pt}
\end{figure}

Real-world robots typically operate in highly structured environments rather than everyday scenarios that contain adversarial objects (e.g., previously unseen or oversized items) and complex constraints (e.g., boxes, shelves, etc.). While a factory robot simply transfers identical items on a belt drive, a domestic robot tasked with rearranging a pantry may frequently encounter oversized containers on shelves. As illustrated in Fig.~\ref{fig:cover}, real-world object rearrangement tasks are inherently heterogeneous, consisting of current objects, goal objects, and environmental constraints. The semantic relationships among these components (e.g., the ``meat can'' on the ``ground'' has a goal location on the ``shelf'') contain essential information for efficiently completing the task. Robots that understand and utilize such knowledge are more likely to succeed in the real world, where adversarial objects and various environmental constraints are ubiquitous.

The object rearrangement problem has traditionally been addressed with model-based task and motion planning (TAMP)~\cite{tamp2021garrett}, which often assumes a fully observable environment and is thus difficult to scale to previously unseen scenarios~\cite{hierachical, wang2021uniform, gao2022bounded}. Recent deep learning-based approaches can generalize to novel objects, owing to the advances in perception and grasping models~\cite{qureshi2021nerp, curtis2022long}. However, they typically assume no environmental constraints (e.g., an open tabletop)~\cite{pan2021decision, tang2022selective, gao2022bounded} or rely on iterative collision checking~\cite{wang2022efficient}, which limits their generalizability in the real world. Additionally, most existing works focus on graspable objects and separately study pick-place or pushing. Although some have investigated both~\cite{pan2021decision, tang2022selective}, they only push to facilitate grasping~\cite{tang2022selective} or employ specialized tools~\cite{pan2021decision}. Relatively few have explored coordinating low-cost pushing, which may be limited by the environment, with pick-place to improve the robot's capability and efficiency. Therefore, the problem of rearranging adversarial objects with multiple skills in constrained environments remains unsolved.

To address this challenge, we propose to learn from the heterogeneous task components and exploit the distinct semantic relationships among them. We devise an adversarial object rearrangement system that utilizes both pushing and grasping, to maximize the robot's efficiency and generalize to novel constrained environments. Our hierarchical approach represents a task as a heterogeneous graph over a pair of current and goal RGB-D images, which are segmented into objects and environmental constraints. At the high-level, a heterogeneous graph neural network~\cite{hetgnn2019} (HetGNN)-based coordinator reasons about the graph and its underlying semantic information and predicts the optimal action primitive and next target, such that the goal configuration can be successfully achieved by the low-level actors with minimal planning costs. The system operates in a closed-loop fashion, continually re-observing the scene at each time step to predict more accurate rearrangement plans.

We experiment in both simulated and real-world environments. Our approach achieves, on average, an 88.78\% success rate with 12.82 actions in real-world tests, outperforming several baselines by large margins. To the best of our knowledge, this is the first approach that utilizes HetGNN to coordinate robot skills for rearranging adversarial objects in constrained environments. The main contributions of this paper are as follows:

\begin{itemize}[leftmargin=*]
\item We propose a hierarchical pushing and grasping robotic system that addresses adversarial object rearrangement problems in constrained environments. By leveraging the semantic relationships in the task, the high-level coordinator guides the 3D CNN-based low-level actors to perform more efficiently.
\item Our approach represents the rearrangement task as a heterogeneous graph and exploits the power of a HetGNN to reason about the underlying relationships among the task components. It learns from an expert planner in simulation and predicts the next target and action end-to-end.  
\item While previous approaches often assume an open workspace or use hard-coded solutions, our method learns to adapt to complex environments, where previously unseen constraints could significantly limit existing works. 
\end{itemize}

\begin{figure*}[t]
    \includegraphics[width=1.0\textwidth]{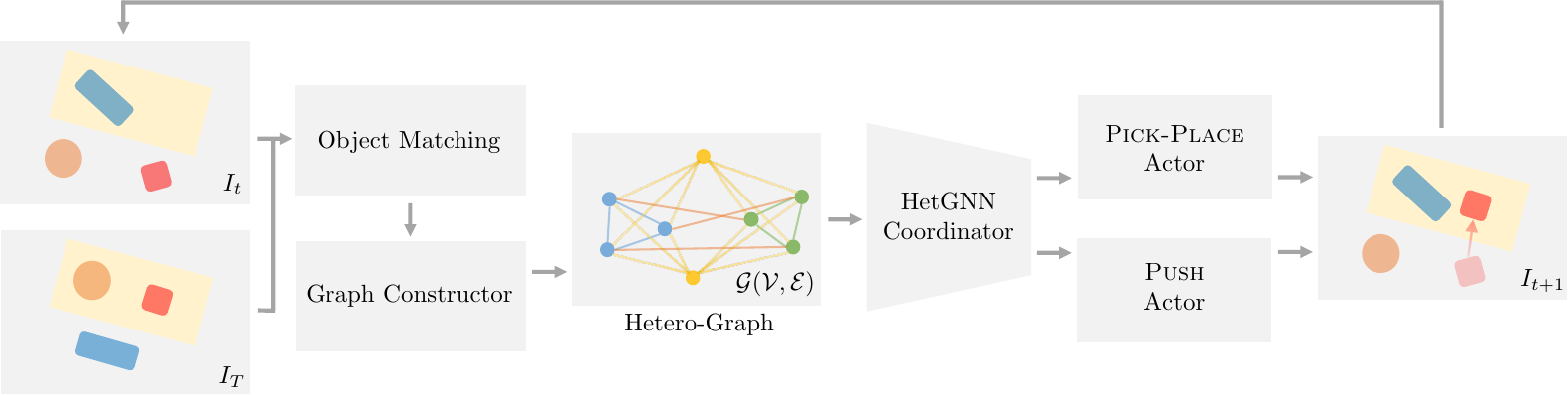}
    \caption{The current RGB-D image $I_t$ and goal $I_T$ are fed into the graph constructor, which encodes the heterogeneous task components (color-coded) into node embeddings with pre-trained 3D encoders. Then the HetGNN updates the embeddings based on its learned parameters, and the high-level coordinator predicts the object selection score $p_o$ and the action selection score $p_{a}$ for each object. We select the object with the highest $p_o$ as the target and decide which action to execute based on $p_{a}$. Finally, we feed the decision to the low-level actors, which are responsible for performing the robot's actions. The closed-loop system will run until the goal configuration is achieved or the maximum number of steps is reached.}
    \label{fig:flowchart}
\vspace{-8pt}
\end{figure*}

\section{RELATED WORK}
% Model-based -> Deep Learning-based: limited to graspable objects
Object rearrangement is an essential challenge in robotics and embodied AI~\cite{batra2020rearrangement}. The problem is commonly studied under the broad subject of task and motion planning (TAMP)~\cite{tamp2021garrett}, which is often formulated hierarchically with a high-level task planner (i.e., which action to perform on which item) and a low-level motion planner (i.e., how to move the end-effector) such that the goals can be achieved~\cite{hierachical, batra2020rearrangement}. Typical TAMP approaches are model-based and often rely on task-specific knowledge and accurate 3D models of the environment~\cite{hierachical, wang2021uniform, gao2022bounded}. Hence, they often do not generalize well to the real world, where the required information may not be accessible. 

Recent works have equipped classical TAMP with deep learning-based perception~\cite{xie2019uois, yu2022self} and grasping models~\cite{lou2020learning, murali2020taskgrasp, sundermeyer2021contact} to generalize to novel objects~\cite{curtis2022long, qureshi2021nerp, Goyal2022IFOR, tang2022selective}. However, many researchers focus exclusively on pick-place~\cite{qureshi2021nerp, pmlr-v164-paxton22a, structformer2022, Goyal2022IFOR}, largely limiting the robot's capability in the real world where objects are frequently not graspable (e.g., large items with a parallel-jaw gripper or cloths with a suction gripper). To rearrange more adversarial objects, non-prehensile action primitives such as pushing are needed. Inspired by~\cite{zeng2018learning}, Tang et al.~\cite{tang2022selective} use pushing to facilitate grasping by breaking the clutter, but not to rearrange adversarial objects. While~\cite{pan2021decision} sorts large-scale basic cuboids with both pushing and grasping, they build a specialized end of arm tooling (EOAT) for pushing. Transporter~\cite{zeng2020transporter} and TRLB~\cite{gao2022bounded} bypass the challenge with suction mechanisms. 
% These advanced hardware makes their problem fundamentally different from ours.

% Semantic info with transformer -> Image-based: we use simple inputs% Rearrangement planning: MCTS, Transformer, GNN -> HGNN we learn better
Long-horizon planning for object rearrangement has been studied analytically with Rapidly Exploring Random Tree (RRT)~\cite{king2016rearrangement} or Monte Carlo Tree Search (MCTS)~\cite{labbe2020}, which explores multiple future possibilities but is less robust to noise and occlusions. PlaNet~\cite{pmlr-v97-hafner19a} addresses the partial observability issue with a learned forward dynamics model and plans actions in latent space. Similarly, Visual Robot Task Planning~\cite{paxton2019visual} learns to encode the scene into a latent representation and then uses tree search for planning in this latent space. Both works are task-specific in simulation and highly likely require large demonstration data to generalize to a real robot. Other researchers have leveraged spatial relations for planning~\cite{pmlr-v164-paxton22a, structformer2022}. Liu et al.~\cite{structformer2022} take language as an input that specifies the goal configuration and then employ Transformers~\cite{vaswani2017attention} to translate the spatial relations into a sequence of pick-place instructions. Our approach conveniently uses a single imperfect RGB-D image to specify the goal and directly transfer to the real world. 

% GNN
Prior robotics research has investigated Graph Neural Networks~\cite{kipf2016gcn, sanchez2020learning} in object rearrangement problems~\cite{wilson20a, qureshi2021nerp, tang2022selective}. Closely related to our work, NeRP~\cite{qureshi2021nerp} employs a high-level object selection module with k-GNNs that plan for rearranging novel objects with pick-place. However, they are limited to graspable objects on an open tabletop, not considering any environmental constraints. Tang et al.~\cite{tang2022selective} compare the Graph Edit Distance (GED) between the start and goal scene graphs and plan for selective object rearrangement of multiple objects, but also assume a simplified environment. In constrained environments, existing works typically assume a constant structure~\cite{cheong2020where} and rely on iterative collision checking~\cite{wang2022efficient}, which is computationally expensive and often suffers from noise and occlusion in the real world. These methods are not as generalizable as ours, as we employ a novel HetGNN-based~\cite{hetgnn2019} coordinator that exploits the semantic relationships among heterogeneous components in the task and significantly improves the robot's efficiency.

\section{PROBLEM FORMULATION}
\label{section:problem}

We aim to design an efficient robotic manipulation system that addresses adversarial object rearrangement problems in unstructured real-world environments, where environmental constraints could heavily influence the robot's behavior. We formulate the problem as follows:

\begin{definition}
\label{def1}
Given a goal image $I_T$ describing a desired object configuration in a constrained environment, the goal of the rearrangement task is to apply a sequence of manipulation actions on the current objects to achieve the goal configuration where every object is within $\tau$ of its corresponding goal location in 3D space.
\end{definition}
In our experiments, we use $\tau = 3 \ cm$.
The constrained environments considered in this work are defined as
\begin{definition}
\label{def2}
Constrained environments include geometric constraints such that certain manipulation actions are not always feasible (e.g., pushing an object across height discontinuities).
\end{definition}

We assume about robot skills and objects as follows:
\begin{assumption}
\label{asm1}
The robot is capable of \textsc{pick}, \textsc{place}, \textsc{move}, and \textsc{push}. \textsc{pick-place} is a sequence of \textsc{pick}, \textsc{move}, and \textsc{place}, while \textsc{push} requires a single \textsc{move}.
\end{assumption}

\begin{assumption}
The adversarial objects are possibly unknown (i.e., novel objects) to the robot and may not be graspable (e.g., object dimension is larger than the maximum opening of the robot's end effector) for which only \textsc{push} is applicable.
\end{assumption}

Let $\mathcal{O}^t = \{o_1^t, o_2^t, \cdots, o_N^t\}$ and $\mathcal{O}^T = \{ o_1^T, o_2^T, \cdots, o_N^T\}$ denote the set of objects in the current scene and the goal scene, respectively. The robot action $a \in \{\textsc{pick-place}, \textsc{push}\}$ for a selected object $o_i^t \in \mathcal{O}^t$ is subject to a binary-valued metric $\mathcal{S}_a(o_i^t, o_i^T, \mathcal{C})\in$ \{0,1\} where $\mathcal{C} = \{c_1, c_2, \cdots, c_N\}$ denotes the set of environmental constraints in the scene. $\mathcal{S}_a = 1$ indicates that \textsc{push} is more effective for the selected object at time $t$, whereas $\mathcal{S}_a = 0$ represents that \textsc{pick-place} is more effective. When both actions are applicable, the robot performs \textsc{push} (i.e., $\mathcal{S}_a = 1$) since it costs fewer actions. 

To reason about the relationships between the objects $\mathcal{O}^t, \mathcal{O}^T$, and the constrained environment $\mathcal{C}$, we employ a heterogeneous graph representation $\mathcal{G}$. 
% \begin{definition}
% \label{def2}
% The heterogeneous graph includes the geometric features of objects in current and goal configurations, the environmental structure information, and the spatial relationships among these components.
% \end{definition}
%
A heterogeneous graph $\mathcal{G}(\mathcal{V, E})$, where $\mathcal{V}$ and $\mathcal{E}$ represent the set of nodes and edges, respectively, is associated with node and edge type mapping functions $\phi \colon \mathcal{V} \rightarrow \mathcal{F}$ and $\psi \colon \mathcal{E} \rightarrow \mathcal{R}$, where $\mathcal{F}$ is the set of node types (e.g., ``current'', ``goal'', and ``environment'') and $\mathcal{R}$ is the set of edge types describing spatial relations (e.g., a ``goal'' node is ``in'' a ``box''). The heterogeneous graph $\mathcal{G}$ is constructed from a pair of RGB-D observations of current and goal configurations $(I_t, I_T)$. We would like to learn a high-level coordinator that predicts a selection probability $p_o(\mathcal{C}, \mathcal{O}^t, \mathcal{O}^T)$ for each object such that the goal configuration can be achieved with the least number of actions by rearranging the most feasible object. The coordinator should simultaneously learn to select the appropriate action for such targets. Specifically, the action probability $p_a(\mathcal{C}, \mathcal{O}^t, \mathcal{O}^T) = p_{push}(\mathcal{C}, \mathcal{O}^t, \mathcal{O}^T) = Pr(\mathcal{S}_a = 1|\mathcal{G(V, E)})$; hence the \textsc{pick-place} probability $p_{pick}(\mathcal{C}, \mathcal{O}^t,\mathcal{O}^T) = Pr(\mathcal{S}_a = 0|\mathcal{G(V, E)}) = 1 - p_{push}$.

\section{PROPOSED APPROACH}
\label{section:proposed_approach}

This section describes the proposed adversarial object rearrangement system that coordinates \textsc{pick-place} and \textsc{push} in constrained environments. To address the exploration challenge in long-horizon problems, our approach takes advantage of the hierarchical structure and uses a high-level coordinator in conjunction with low-level actors to guide the robot at each time step $t$. The goal configuration at time $T$ is given as a reference RGB-D image $I_T$. Given the current observation of the scene $I_t$, the HetGNN-based coordinator reasons about the underlying relationships in the heterogeneous graph and simultaneously predicts which object should be prioritized and how to move it, such that the goal can be achieved efficiently. The overview of the approach is described in Fig.~\ref{fig:flowchart}, and the algorithm is delineated in Algorithm~\ref{algo:rearrangement}.
% The algorithm is delineated in Algorithm~\ref{algo:rearrangement} and the following subsections explain it in detail.

\begin{algorithm}[t]
\caption{Rearrangement in Constrained Environments}
\label{algo:rearrangement}
\hspace*{\algorithmicindent} \textbf{Input: } Goal Image $I_T$, encoder $E_{\phi}$, Coordinator $\mathcal{N}_c$\\
\hspace*{\algorithmicindent} \textbf{Output: } $o^t \in \mathcal{O}^t, a \in \mathcal{A}$
\begin{algorithmic}[1]
\State $\mathcal{M}_T, \mathcal{M}_C \gets \texttt{Segmentation($I_T$)}$
\State $N \gets \texttt{length($\mathcal{M}_T$)}$
\State $\mathcal{H}_T \gets\texttt{ExtractFeature($\mathcal{M}_T, I_T$)}$
\State $\mathcal{P}_T, \mathcal{P}_c, \mathbf{z}_T, \mathbf{z}_c \gets \texttt{BackProjection($\mathcal{M}_T, \mathcal{M}_c, I_T$)}$
\State $V_T, V_c \gets \texttt{VoxelTransformation}(\mathcal{P}_T, \mathcal{P}_c)$
\State $\mathbf{x}_T, \mathbf{x}_c \gets E_{\phi}(V_T, V_c) \cup (\mathbf{z}_T, \mathbf{z}_c)$
\For{${t}\in\ {0, ..., T-1}$}
    \State $I_t \gets \texttt{Observation($\mathcal{W}$)}$
    \State $\mathcal{M} \gets \texttt{Segmentation($I_t$)}$
    \State $\mathcal{H}_t \gets \texttt{ExtractFeature($\mathcal{M}, I_t$)}$
    \State $\mathbf{c} \gets \texttt{Match($\mathcal{H}_t, \mathcal{H}_T, N$)}$
    \State $\mathcal{P}_t, \mathbf{z}_t \gets \texttt{BackProjection($\mathcal{M}, I_t$)}$
    \State $V_t \gets \texttt{VoxelTransformation}(\mathcal{P}_t)$
    \State $\mathbf{x}_t \gets E_{\phi}(V_t) \cup \mathbf{z}_t$
    \State $\mathcal{G(V, E)} \gets \texttt{BuildGraph($\mathbf{x}_{t}, \mathbf{x}_{T}, \mathbf{x}_c, \mathbf{c}$)}$
    \State $p_a, p_o \gets \texttt{$\mathcal{N}_c$.Feedforward}(\mathcal{G})$
    \State $o^t \gets \argmax_{o^t \in \mathcal{O}^t} p_o(\mathcal{C}, \mathcal{O}^t, \mathcal{O}^T)$
    \If{$p_a > 0.5$}
        \State $a \gets \texttt{Push}(o^t)$
    \Else
        \State $a \gets \texttt{Pick-place}(o^t)$
    \EndIf
    \State $\texttt{Actors}(a)$
\EndFor
\end{algorithmic}
\end{algorithm}

\subsection{Object Matching}
\label{section:matching}
Given the goal configuration specified by an RGB-D image $I_T$, the object matching module finds each object’s correspondence in the current observation $I_t$. We first obtain the instance masks $\mathcal{M}_T$ of $N$ objects in $I_T$ using the SAG~\cite{yu2022self}, an object instance segmentation method with active robotic manipulation. Next, we encode each object's RGB-D cropping from $\mathcal{M}_T$ into a feature vector $\mathbf{h}_i \in \mathbb{R}^{10}$ using a Siamese network~\cite{Koch2015SiameseNN}. The network is trained with contrastive loss such that the L2 distance in the latent space is close for the same objects and far for different ones~\cite{danielczuk2019mechanical, lou2021collision}. The set $\mathcal{H}_T= \{ \mathbf{h}_1^T, \mathbf{h}_2^T, \cdots , \mathbf{h}_N^T \}$ represents the features of objects in the goal configuration. At the current time step $t$, we follow the same procedure to extract the feature set $\mathcal{H}_t$ of current objects. The L2 distance between each element in $\mathcal{H}_t$ and each element in $\mathcal{H}_T$ is calculated, and the current-goal correspondence $\mathbf{c} \in \mathbb{R}^{N \times 2}$ is established by associating each goal object to the one with the smallest L2 distance in the current time $t$. 

\subsection{Constructing Heterogeneous Graph}
\label{section:construct_het}

The high-level coordinator is based on a HetGNN, which exploits the heterogeneity and the underlying semantic information in the input heterogeneous graph. To construct a graph that can efficiently capture the information, we consider three different node types: current objects $\mathcal{O}^t$, goal objects $\mathcal{O}^T$, and the environmental constraints $\mathcal{C}$. Unlike the traditional homogeneous graphs, the relationships between these nodes are represented by a set of heterogeneous edge types, which could be semantically interpreted (e.g., the edge between ``current objects'' nodes and ``constraints'' nodes representing the ``in'' relationship, the edge between ``current objects'' nodes and ``goal object'' nodes representing the ``to'' relationship). 

The heterogeneous graph is illustrated in Fig.~\ref{fig:flowchart}. The nodes $\mathcal{V}$ include current nodes $\mathbf{v}^t$, goal nodes $\mathbf{v}^T$, and the constraints nodes $\mathbf{v}^c$, representing different types of the heterogeneous task components. The graph connectivity contains two fully-connected sub-graphs, one for current nodes and one for goals nodes. Each current node is also connected to its corresponding goal node, specified by the current-goal correspondence $\mathbf{c}$. The constraint node(s) is/are individually connected to each object node to propagate the influence of the environmental constraints. Each node embedding is extracted from the geometric shape of the object or environment. Specifically, the point clouds of the current objects $\mathcal{P}_t$, goal objects $\mathcal{P}_T$, and constraints $\mathcal{P}_c$ are obtained through back-projection, and transformed into voxel grids $V_t$, $V_T$, and $V_c$, respectively. We then encode the voxel grids into geometric features $\mathbf{x}_t$, $\mathbf{x}_T$, and $\mathbf{x}_c$ using a 3D encoder $E_{\phi}$: Conv3D(1, 32, 5) $\rightarrow$ ELU $\rightarrow$ Maxpool(2) $\rightarrow$ Conv3D(32, 32, 3) $\rightarrow$ ELU $\rightarrow$ Maxpool(2) $\rightarrow$ FC($32 \times 6 \times 6 \times 6$, 12). The encoder is taken from a pretrained 3D Convolutional Autoencoder, whose latent features could effectively represent the shape of the input object. Finally, each node embedding is concatenated with the object's location $\mathbf{z} \in \mathbb{R}^3$.

\subsection{HetGNN-based Coordinator} 

\begin{figure}[h]
    \includegraphics[width=1.0\linewidth]{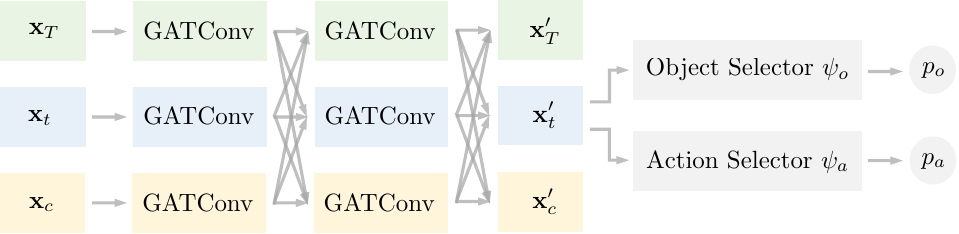}
    \caption{The HetGNN network takes as input graphs of heterogeneous node types (e.g., $\mathbf{x}_T, \mathbf{x}_t, \mathbf{x}_c$). The message passing functions are duplicated for each edge type to update the weights for different relationships. Finally, the scores $p_a$ and $p_o$ are derived from updated current node features $\mathbf{x}'_T, \mathbf{x}'_t, \mathbf{x}'_c$.}
    \label{fig:hetgnn}
\vspace{-8pt}
\end{figure}

GNNs are effective in discovering underlying relationships among nodes by learning a non-linear function $\mathcal{F}$, which encodes a graph $\mathcal{G}$ to $\mathcal{G}'$ with updated node and edge features~\cite{zhou2020graph}. 
% The GNN combining multiple layers $l$ of graph encoders are described by
% \begin{align}
% \label{eq:GNN}
% \mathcal{G}^{(l+1)} = \mathcal{F}^{(l)} \left( \mathrm{ReLU} \left( \mathcal{G}^{(l)}(\mathcal{V}^{(l)}, \mathcal{E}^{(l)}) \right) \right), \forall l \in [0, L].
% \end{align}
We start with the base homogeneous Graph Attention Networks (GAT)~\cite{velickovic2018graph}. The message-passing function, parameterized by a weight matrix $\mathbf{\Theta}$ and attention coefficients $\alpha_{i, j}$, for updating latent features $\mathbf{x}_i$ of node $\mathbf{v}_i$ is defined as
\begin{align}
\label{eq:GAT1}
\mathbf{x}^{\prime}_i = \alpha_{i,i}\mathbf{\Theta}\mathbf{x}_{i} +
\sum_{j \in \mathcal{N}(i)} \alpha_{i,j}\mathbf{\Theta}\mathbf{x}_{j}
\end{align}
where the attention coefficients $\alpha_{i, j}$ are computed by
\begin{align}
\label{eq:GAT2}
\alpha_{i,j} =
\frac{
\exp\left(\mathrm{\sigma}\left(\mathbf{a}^{\top}
[\mathbf{\Theta}\mathbf{x}_i \, \Vert \, \mathbf{\Theta}\mathbf{x}_j]
\right)\right)}
{\sum_{k \in \mathcal{N}(i) \cup \{ i \}}
\exp\left(\mathrm{\sigma}\left(\mathbf{a}^{\top}
[\mathbf{\Theta}\mathbf{x}_i \, \Vert \, \mathbf{\Theta}\mathbf{x}_k]
\right)\right)}
\end{align}
The $\mathbf{a}$ is the learned weight matrix of the attention mechanism, $\mathcal{N}(i)$ is the neighbors of $\mathbf{v}_i$, and $\sigma = LeakyReLU(\cdot)$.

Note that homogeneous graph neural networks could not differentiate different types of nodes and edges. They lack the mechanism to effectively harness the heterogeneous information. To exploit the semantic relationships among the heterogeneous task components, we adopt the approach in~\cite{hetgnn2019} that introduces heterogeneity to the homogeneous GNN by dedicating an individual message passing function to each edge type, as shown in Fig.~\ref{fig:hetgnn}. Given a heterogeneous graph $\mathcal{G(V, E)}$, the network aggregates node embeddings by using the message passing functions corresponding to the active edge types, which are determined by the types of the connected nodes. For instance, the edge between current nodes $\mathbf{v}_t$ and goal nodes $\mathbf{v}_T$ belongs to a ``current-to-goal'' edge type. The HetGNN includes three graph attention convolutional layers to ensure effective learning of the underlying relational information. After the node embeddings are updated by the HetGNN, two Multi-Layer-Perceptron (MLP)-based prediction heads, object selector $\psi_o: \mathcal{V} \rightarrow \mathcal{O}$ and action selector $\psi_a: \mathcal{V} \rightarrow \mathcal{A}$, are connected to $\mathbf{x}_t$ to estimate which action should be performed on which object.

\subsection{Low-level Actors}
The low-level actors are responsible for executing the actions decided by the high-level coordinator. If \textsc{pick-place} is selected, we first generate a batch of grasp candidates following the shape completion-based sampling algorithm in~\cite{lou2022grasping}. Next, we use the Grasp Stability Predictor (GSP)\footnote{Note that the acronym GSP refers to the 3D CNN grasping module in~\cite{lou2020learning}, defined here for a concise reference.}, a 3D CNN-based 6-DoF grasp detection algorithm, to select a feasible pose for grasping. After the target object has been successfully grasped, we place the object to its corresponding goal location by checking the current-goal correspondence $\mathbf{c}$ calculated in~\ref{section:matching}. If \textsc{push} is the more effective action for the target, we plan for a direct pushing path while checking collisions using the flexible collision library (FCL)~\cite{pan2012fcl}. The robot closes its fingers and follows a straight path, which is divided into multiple short segments of fixed length by the intermediate waypoints. Then we use the mean square error (MSE) between the object's voxel grids and goal location to supervise a simplified model predictive control loop.

\subsection{Expert Planner and Training}
\label{section:data_collection}
\begin{figure}[t]
    \includegraphics[width=0.485\textwidth]{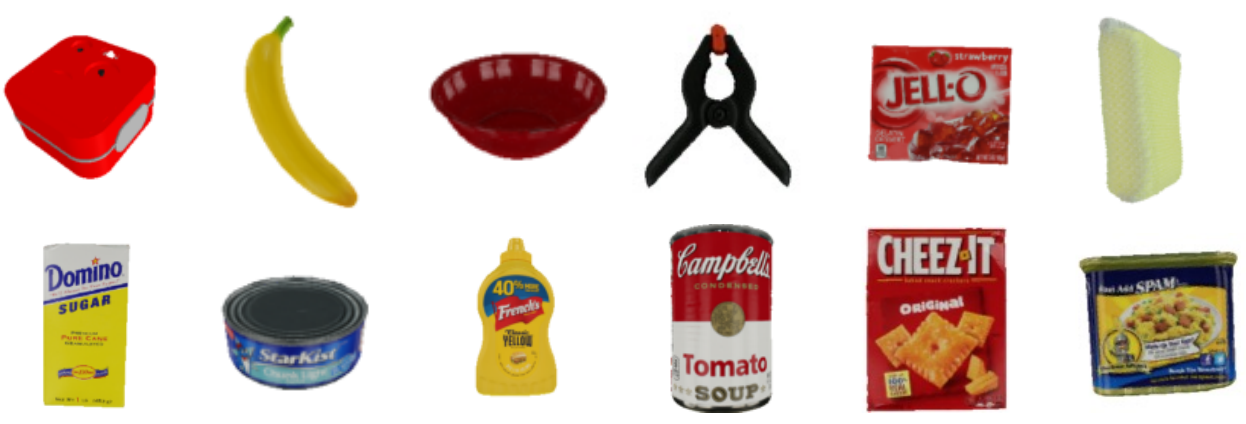}
    \caption{The testing objects are drawn from the YCB dataset and differ in size, color, and shape from our basic training objects (e.g., blocks, cylinders). Some objects such as bowls and cracker boxes, may not be graspable due to their orientation.}
  \label{fig:training_objects}
  \vspace{-8pt}
\end{figure}

To obtain training data, we generated 3,000 RGB-D images of the randomly positioned training objects (e.g., toy blocks and cylinders of different size) in environments with arbitrary constraints (e.g., bins, shelves). Then, two RGB-D images are randomly sampled as start and goal configurations for a rearrangement task. For all the rearrangement tasks, we define a pick-place cost of \textbf{3} and a push cost of \textbf{1} following Assumption~\ref{asm1}.

The training labels, action selection labels, and object selection labels are automatically annotated by an expert planner built in a fully observable simulator. First, the expert planner examines each object's mesh model and reasons about which action primitive to use. It relies on two criteria: 1) if the object is graspable and 2) if a direct pushing path exists between the current and goal location. The binary action selection label will be 1 for \textsc{push} if the object is not graspable or a direct pushing path exists, and 0 for \textsc{pick-place} if the object is graspable and no direct pushing path exists. We assume there are no invalid tasks such as moving an ungraspable object across a discontinuous path (e.g., moving a large plate from the table onto the shelf). Then, based on the action assigned to the objects, the expert planner computes the optimal planning solution analytically using the $A^\ast$ algorithm, which globally minimizes the predefined operating cost function by computing all possible planning sequences. For an ungraspable object without a direct pushing path, we assign an action cost of 3 because it requires multiple pushing actions. An infinite heuristic cost is associated with a planning sequence if the goal location is blocked by other objects, determined by FCL. The binary object selection label is 1 if it is the first in a sequence planned by $A^\ast$ and 0 otherwise. The dataset contains 30,000 pairs of current and goal RGB-D images and labels, which are transformed into heterogeneous graphs using methods described in Sec.~\ref{section:construct_het}.

During training, we use a binary cross-entropy loss $\mathcal{L}_{action}$ to supervise the action predictions:
\begin{align}
    \label{eq:bce}
    \mathcal{L}_{action} = -{(y\log(p) + (1 - y)\log(1 - p))}.
\end{align}
The Huber loss $\mathcal{L}_{object}$ for the object prediction head output $\hat{y}$ is defined as
\begin{align}
    \label{eq:action}
    \mathcal{L}_{object}=
        \left\{\begin{matrix}
            \frac{1}{2}(y - \hat{y})^{2} & \text{if}~\left | (y - \hat{y})  \right | < \delta\\
            \delta ((y - \hat{y}) - \frac1 2 \delta) & \text{otherwise}
    \end{matrix}\right.\end{align}
where $\delta = 1.15$.
The combined loss $\mathcal{L}$ is defined as
\begin{align}
    \mathcal{L} = \mathcal{L}_{object} + \lambda \mathcal{L}_{action}.
\end{align}
We empirically found that $\lambda = 0.65$ yields the best performance for our problem.

\section{EXPERIMENTS}

We experiment in both simulated and real-world settings. These experiments are designed to: 1) demonstrate the effectiveness of our hierarchical system for the adversarial object rearrangement problem; 2) evaluate our HetGNN-based coordinator in various constrained environments and compare it to other baselines; and 3) show the generalizability of our approach to unstructured everyday scenarios.

\textbf{Evaluation metrics}: Following Definition~\ref{def1}, we define the \textit{success rate} as $ \frac{\text{\# of successful rearrangement}}{\text{\# of total rearrangement problems}}$. If a given rearrangement is not achievable (e.g., lifting a non-graspable object from the ground to the upper shelf), the experiment will be re-initialized. Each test is limited to $2 \times N$ planning steps, where $N$ is the number of objects. A timeout is also considered a failure. We also consider the \textit{planning cost}, which measures the number of actions taken to rearrange the objects from the start to the goal configuration. Each \textsc{push} costs \textbf{1} action and each \textsc{pick-place} costs \textbf{3} actions by Definition~\ref{def2}. Because \textsc{pick-place} only approaches could not work in our settings due to the non-graspable objects, we instead compare our method with the following four baselines: 
\begin{itemize}
    % Model-based approach as in NeRP, suggested by the reviewer
    \item \textbf{\texttt{Model}} is a model-based approach that assumes access to ground truth IDs and mesh models. It randomly selects a target and checks if its corresponding goal location is available. If that location is occupied, it will push the occupying object to an arbitrary free space. Otherwise, it moves the object using the expert action selection algorithm described in Sec.~\ref{section:data_collection}.
    \item \textbf{\texttt{Plan}} is a variant of the expert planner in Sec.~\ref{section:data_collection}. It combines an optimal planner with a deep learning-based perception module~\cite{yu2022self}. Instead of using 3D models, this classical approach is based-on segmentation masks and plans for the entire action sequence with $A^\ast$ algorithm that globally minimizes the cost function.
    \item \textbf{\texttt{GNN}} employs a Homogeneous Graph Neural Network instead of HetGNN for the coordinator. The network is trained with the same dataset as ours, except that the heterogeneous structure is not used. All other components are kept the same as in our approach.
    \item \textbf{\texttt{NeRP+Push}} builds upon a recent state-of-the-art object rearrangement approach~\cite{qureshi2021nerp}. It learns a k-GNN-based planner that selects a near-optimal policy and uses \textsc{pick-place} for unknown object rearrangement. To adapt to our test environments, we allow NeRP to heuristically push ungraspable objects when the object mask is larger than a threshold.
    \item \textbf{{\texttt{Expert}}} is our expert planner in Sec.~\ref{section:data_collection} whose performance is regarded as the upper bound of each scenario. It computes the optimal solution but may fail due to unexpected object dynamics and imperfect low-level actors.
\end{itemize}

\begin{figure}[t]
    \includegraphics[width=0.485\textwidth]{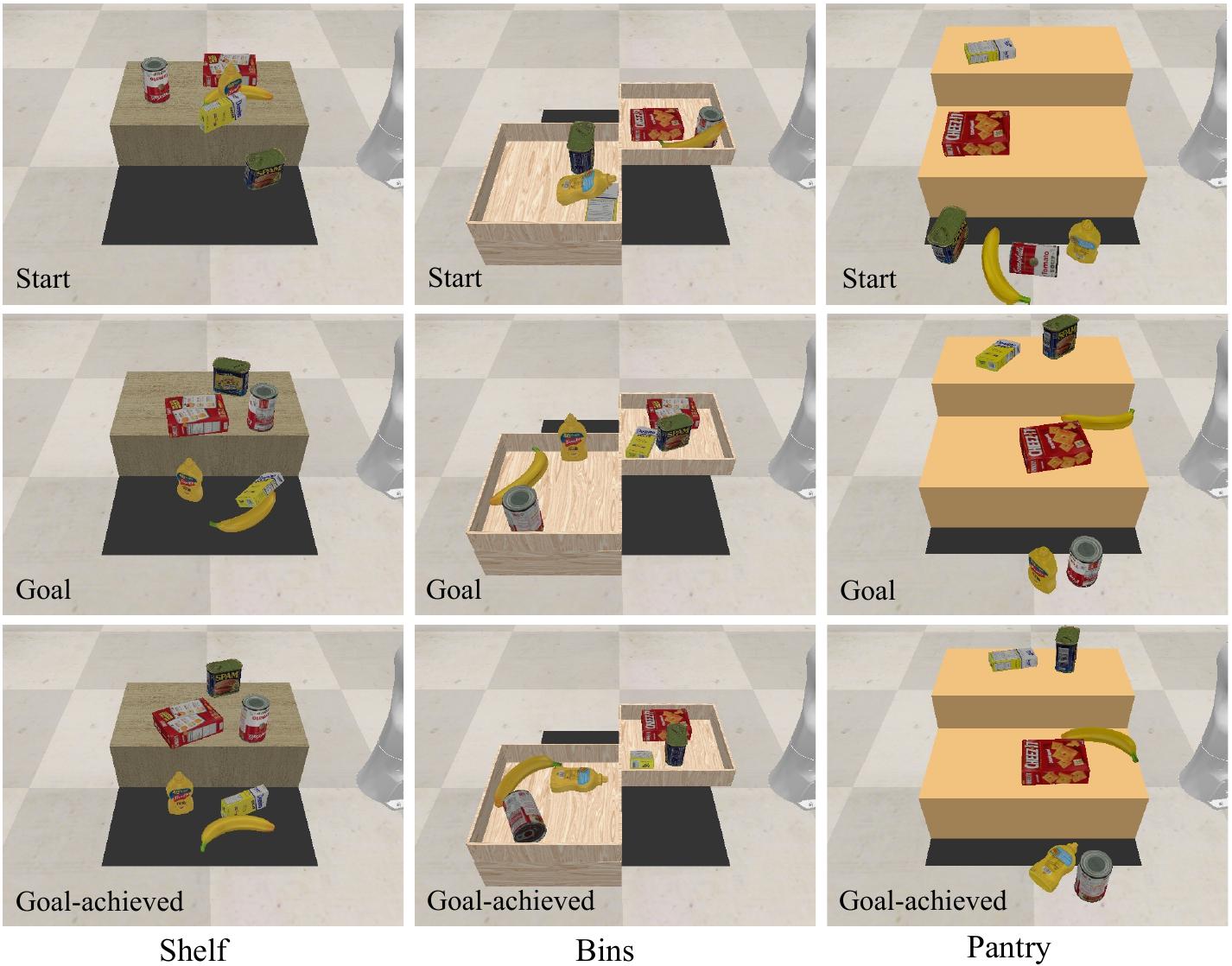}
    \caption{The simulated experiments include various constraints that increase clutteredness and planning difficulty.  Our approach achieves comparable results to the expert planner and outperforms all the baseline methods.}
  \label{fig:sim_exp}
  \vspace{-8pt}
\end{figure}

\begin{table}[t]
\caption{Success Rate (\%) on Tabletop}
\vspace{-8pt}
\label{tab:sim_result1_a}
\begin{center}
\begin{tabular}{ccccccc}
\toprule
  & Model & Plan & GNN & NeRP+Push & Ours & Expert\\
\midrule
5-object & 94.12 & \textbf{98.04} & 88.24 & 94.12 & \textbf{98.04} & 100.00\\
6-object & 90.02 & 92.16 & 90.20 & \textbf{98.04} & \textbf{98.04} & 100.00\\
7-object & 84.27 & 88.24 & 82.35 & 90.20 & \textbf{94.12} & 100.00\\
\bottomrule
\end{tabular}
\end{center}

\caption{Planning Cost (\# actions) on Tabletop}
\vspace{-8pt}
\label{tab:sim_result1_b}
\begin{center}
\begin{tabular}{ccccccc}
\toprule
  & Model & Plan & GNN & NeRP+Push & Ours & Expert\\
\midrule
5-object & 12.33 & 10.12 & 13.49 & 19.41 & \textbf{9.67} & 8.42\\
6-object & 15.42 & 15.47 & 16.69 & 22.13 & \textbf{11.91} & 10.12\\
7-object & 18.96 & 18.41 & 26.58 & 24.27 & \textbf{15.60} & 13.54\\
\bottomrule
\end{tabular}
\end{center}
\vspace{-8pt}
\end{table}

\begin{table}[t]
\caption{Success Rate (\%) in Constraints}
\vspace{-8pt}
\label{tab:sim_result2_a}
\begin{center}
\begin{tabular}{ccccccc}
\toprule
  & Model & Plan & GNN & NeRP+Push & Ours & Expert\\
\midrule
Shelf & 88.24 & 96.08 & 82.36 & \textbf{98.04} & \textbf{98.04} & 100.00\\  
Bins & 82.35 & 78.43 & 72.55 & 84.31 & \textbf{98.04} & 100.00\\
Pantry & 84.31 & 82.35 & 72.55 & 88.24 & \textbf{94.12} & 100.00\\
\bottomrule
\end{tabular}
\end{center}
% \end{table}
% \begin{table}[b]
\caption{Planning Cost (\# actions) in Constraints}
\vspace{-8pt}
\label{tab:sim_result2_b}
\begin{center}
\begin{tabular}{ccccccc}
\toprule
  & Model & Plan & GNN & NeRP+Push & Ours & Expert\\
\midrule
Shelf & 16.64 & 16.12 & 17.58 & 21.56 & \textbf{14.19} & 13.82\\
Bins & 20.56 & 18.08 & 18.91 & 24.83 & \textbf{15.64} & 14.21\\
Pantry & 18.16 & 17.45 & 19.83 & 22.21 & \textbf{14.68} & 13.65\\
\bottomrule
\end{tabular}
\end{center}
\end{table}

\begin{figure*}[t]
    \includegraphics[width=1.0\textwidth]{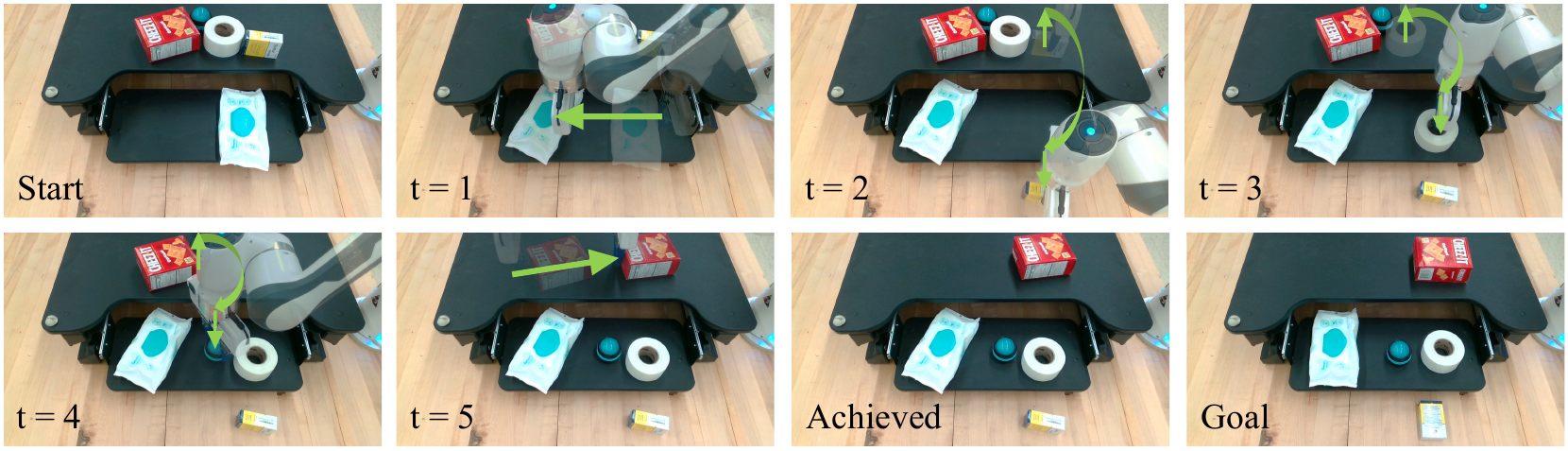}
    \caption{An example of real-world experiments in the \textit{Novel} scenario (11 actions). The HetGNN-based coordinator predicts the most feasible target and utilizes \textsc{pick-place} and \textsc{push} accordingly while the low-level actors execute the plan in closed-loop.}
\label{fig:real_exp}
\vspace{-8pt}
\end{figure*}
\subsection{Simulation Experiments}

The simulated test environment is in CoppeliaSim 4.0 ~\cite{rohmer2013v} with Bullet physics engine v2.83. The scene includes a Franka Emika Panda robot arm with the original gripper, different numbers of testing objects, and various environmental constraints. A single-view RGB-D observation is taken with a simulated Kinect camera. 

\textbf{Experiment scenes:} The experiments are depicted in Fig.~\ref{fig:sim_exp}. We first test in the open \textit{Tabletop} scenario for which the baseline methods were designed to verify if our approach is efficient in the simple environment. Each scene contains five to seven objects from Fig.~\ref{fig:training_objects} to demonstrate that our approach is generalizable to different numbers of objects (i.e., clutteredness). Each method is tested 51 times, and the success rate and planning cost are compiled in Table~\ref{tab:sim_result1_a} and~\ref{tab:sim_result1_b} respectively. Then we experiment with increasing the complexity of the environmental constraints: \textit{Shelf} demonstrates that our approach is able to efficiently solve multi-planar scenarios; \textit{Bins} is commonly seen in warehouses and introduces more partial occlusions; the additional shelves in \textit{Pantry} mimic a more realistic scene and show the generalizability to novel constraints, where analytical approaches often face difficulties. The experiments contain six YCB objects, and the results are compiled in Table~\ref{tab:sim_result2_a} and ~\ref{tab:sim_result2_b}.

\textbf{Result Analysis:} In the \textit{Tabletop} experiments, our approach achieved a 94.12\% success rate with a 15.60 planning cost in the most cluttered \textit{7-object} scenario, surpassing all baseline methods. The major challenge is to understand the relations between current and goal objects. Although \textit{Model} has access to all the mesh models and the expert action selection module, it performs poorly owing to the uninformed planner, which frequently proposes targets whose goal locations are occupied. \textit{Plan}'s open-loop planner could not resolve a collision immediately, potentially causing more failures and requiring more actions to complete the task afterward. \textit{GNN} learns less effectively and makes predictions that are not as accurate as ours (e.g., selecting \textsc{pick-place} while \textsc{push} is feasible). \textit{NeRP+Push} only pushes when the item is not graspable, since it could not efficiently coordinate different skills. The results suggest that our approach is able to generalize to different numbers of adversarial objects and is the most efficient in cluttered scenes, thanks to the HetGNN coordinator and the closed-loop design.

Constrained experiments increase environmental discontinuities and clutteredness, necessitating more accurate reasoning about the relationships among task components. The privileged information available to \textit{Model} helps maintain its performance, while \textit{GNN} becomes worse since it has no knowledge of the relations between each component. This indicates that the information learned by HetGNN is crucial to efficient planning. \textit{Plan} depends on accurate object masks to calculate the true trajectory cost, and \textit{NeRP+Push} only considers objects' center locations. Consequently, their performance suffers from the additional challenge of constrained experiments. In contrast, our HetGNN-based coordinator that employs 3D shape features generalizes better with partial observations. Overall, we achieved an average success rate of 96.73\%, surpassing the best-performing baseline by 7.2\%, with a planning cost of only 14.50, which is the closest to \textit{Expert}'s result.

\subsection{Real-robot Experiments}

Our real-world experiment consists of a Franka Emika Panda robot arm with FESTO DHAS soft fingers and an Intel RealSense D415 camera that overlooks the workspace. We test each method 11 times in three real-world scenarios: \textit{Bins}, \textit{Shelf}, and \textit{Novel}. \textit{Model} is excluded from the baseline methods because ground truth mesh models are not available in the real world. Five adversarial objects are first randomly placed in the scene as the goal configuration and then re-initialized to the start configuration. To address the sim-to-real gap of the RGB-D sensor, we fine-tuned the object-matching module with a dataset consisting of 500 real-world images.

Fig.~\ref{fig:real_exp} depicts a successful rearrangement task in novel scenarios. Due to the perception challenge and more complex object dynamics in the real world, the performance of all the methods declines compared to the simulated results. However, ours drops much less thanks to the learned model that is robust to noise and occlusion. \textit{Plan} and \textit{NeRP+Push} sometimes select the wrong action and falsely store objects when the goal locations are available because of inaccurate masks and noise. Collisions with the occluded geometry also become more frequent in the real world, highlighting the importance of our closed-loop system. We summarized the experimental results in Table~\ref{tab:real_results_a} and~\ref{tab:real_results_b}. Our approach achieves an average success rate of 87.88\% and completes the task with 12.82 actions, indicating that it is the most efficient and generalizable to novel office objects and constraints. Our method is limited by the analytical pushing algorithm, which may rotate large objects unexpectedly and incur additional costs if they collide with other objects.

\begin{table}[b]
\caption{Success Rate (\%) in Real World}
\vspace{-8pt}
\label{tab:real_results_a}
\begin{center}
\begin{tabular}{cccccc}
\toprule
  & Plan & GNN & NeRP+Push & Ours\\
\midrule
Shelf & 81.82 & 72.73 & \textbf{90.91} & \textbf{90.91}\\  
Bins & 72.73 & 63.64 & 72.73 & \textbf{81.82}\\
Novel & 63.64 & 72.73 & 81.82 & \textbf{90.91}\\
\bottomrule
\end{tabular}
\end{center}
% \end{table}

% \begin{table}[b]
\caption{Planning Cost (\# actions) in Real World}
\vspace{-8pt}
\label{tab:real_results_b}
\begin{center}
\begin{tabular}{cccccc}
\toprule
  & Plan & GNN & NeRP+Push & Ours\\
\midrule
Shelf & 14.62 & 16.67 & 19.56 & \textbf{11.81}\\
Bins & 19.13 & 17.88 & 23.83 & \textbf{14.34}\\
Novel & 20.20 & 17.37 & 21.21 & \textbf{12.31}\\
\bottomrule
\end{tabular}
\end{center}
\end{table}

% \begin{figure}[t]
%     \includegraphics[width=0.485\textwidth]{images/failed_example.pdf}
%     % \caption{In densely cluttered scenarios, the starting push point could be blocked by the other objects and causes the push actor unable to complete the task.}
%     \caption{Without HetGNN}
%   \label{fig:failures}
%   \vspace{-8pt}
% \end{figure}

\section{CONCLUSION}
We presented an object rearrangement system that coordinates \textsc{pick-place} and \textsc{push} in challenging scenarios with adversarial objects and environmental constraints. Our approach hierarchically employs a HetGNN coordinator and low-level 3D CNN-based actors to achieve the goal arrangement in an efficient manner. The proposed simulation-trained rearrangement system achieved an average success rate of 87.88\% and a planning cost of 12.82 in real-world experiments with adversarial objects and environmental constraints. One avenue for future extension is to simultaneously learn the orientations of objects during placement, as we are currently focusing on arranging objects in terms of positions.

%%%%%%%%%%%%%%%%%%%%%%%%%%%%%%%%%%%%%%%%%%%%%%%%%%%%%%%%%%%%%%%%%%%%%%%%%%%%%%%%
\bibliographystyle{IEEEtran}
\bibliography{IEEEabrv,IEEEexample}
\end{document}